\documentclass[10pt,twocolumn,letterpaper]{article}

\usepackage{wacv}
\usepackage{times}
\usepackage{epsfig}
\usepackage{graphicx}
\usepackage{amsmath}
\usepackage{amssymb}
\usepackage{subcaption}
\usepackage{url}

\usepackage{array}
\usepackage{multirow}
\usepackage{textcomp}
\usepackage{array}
\usepackage{colortbl}
\usepackage[table]{xcolor}
\usepackage{booktabs}
\usepackage{multirow}

\definecolor{myblue}{rgb}{0,0,1}
\definecolor{myred}{rgb}{0.8, 0, 0}
\definecolor{mygreen}{rgb}{0, 0.6, 0}

\wacvfinalcopy 

\def\wacvPaperID{1128} 

\ifwacvfinal\fi
\begin{document}

\title{Post-Mortem Iris Recognition Resistant to Biological Eye Decay Processes}

\author{Mateusz Trokielewicz \\
Research and Academic Computer Network NASK, Warsaw, Poland\\
{\tt\small mateusz.trokielewicz@nask.pl}
\and
Adam Czajka \\
University of Notre Dame, IN, USA\\
{\tt\small aczajka@nd.edu}
\and
Piotr Maciejewicz \\
Medical University of Warsaw, Warsaw, Poland\\
{\tt\small piotr.maciejewicz@wum.edu.pl}
}

\maketitle
\ifwacvfinal\thispagestyle{empty}\fi

\begin{abstract}
This paper proposes an end-to-end iris recognition method designed specifically for post-mortem samples, and thus serving as a perfect application for iris biometrics in forensics. To our knowledge, it is the first method specific for verification of iris samples acquired after demise. We have fine-tuned a convolutional neural network-based segmentation model with a large set of diversified iris data (including post-mortem and diseased eyes), and combined Gabor kernels with newly designed, iris-specific kernels learnt by Siamese networks. The resulting method significantly outperforms the existing off-the-shelf iris recognition methods (both academic and commercial) on the newly collected database of post-mortem iris images and for all available time horizons since death. We make all models and the method itself available along with this paper.
\end{abstract}

\section{Introduction}
Identification of deceased subjects with their iris patterns has recently moved from the science-fiction realm into scientific reality thanks to ongoing research that has been happening in this field for several years now. Despite some popular claims, researchers have shown that post-mortem iris recognition can be viable under favorable conditions, such as cold storage in a hospital morgue \cite{BostonPostMortem, TrokielewiczPostMortemICB2016, TrokielewiczPostMortemBTAS2016, TrokielewiczTIFS2018}, but challenging in outdoor environment, especially in the summer \cite{BolmeBTAS2016, Sauerwein_JFO_2017}. A method for detecting presentation attacks employing cadaver eyes has been also proposed \cite{TrokielewiczColdPAD_BTAS2018}, as well as experiments comparing machine and human attention to iris features in post-mortem iris recognition proceedings \cite{Trokielewicz_BTAS_2019}.

However, an end-to-end iris recognition method designed specifically for forensic applications has not yet been proposed, and all published experiments that assess feasibility of post-mortem iris matching are based on methods designed for samples acquired from live subjects. This may significantly underestimate the true performance due to multiple factors related to post-mortem eye decomposition not considered in off-the-shelf iris matchers. In this paper we make an attempt at improving the efficiency of iris representation by introducing data-driven kernels that are {\bf learnt from post-mortem iris images}. A shallow (incorporating only one convolutional layer, as in dominant Daugman's iris recognition algorithm) Siamese network is employed for learning a novel descriptor in a form of two dimensional filter kernels that can be further used in any conventional iris recognition pipeline. An open-source OSIRIS implementation has been chosen to demonstrate the superiority of the newly designed kernel set, not only over the original approximations of Gabor kernels, but also over an example commercial iris recognition method. In addition, we fine-tuned the convolutional neural network (CNN) designed for post-mortem iris segmentation \cite{TrokielewiczIMAVIS2019arxiv} with more diverse training data, which allowed for further increase in the overall recognition accuracy.

Thus, this paper offers the first known to us and complete method designed specifically for post-mortem iris recognition, with the following {\bf contributions}:

\begin{itemize}
	\item a new, post-mortem iris-specific feature representation method comprising filters learned from post-mortem data, offering a significant reduction of recognition errors when compared to methods designed for live irises,
	\item analysis of false non-match rates at different false match thresholds suitable for a forensic setting, 
	\item an updated CNN-based iris segmentation model, fine-tuned with more diverse iris samples, offering better robustness against unusual deviations from the ISO/IEC 19794-6 observed in post-mortem iris data,
	\item source codes for iris-specific filter training experiments, trained filter kernels, and new CNN iris segmentation model -- everyting that is needed to fully replicate the experiments.  
\end{itemize}

\section{Related Work}
\label{sec:Related}


\subsection{One-shot recognition and Siamese networks}

Recent advancements in deep learning allowed CNN-based image classifiers to achieve performance superior to many hand-crafted methods. However, one of the downsides of deep learning is the need for large quantities of labeled data in case of supervised learning. This becomes a problem in applications where prediction must be obtained about the data belonging to an under-sampled class, or about a class unknown during training.

Siamese networks, on the other hand, perform well in the so called \emph{one-shot} recognition tasks, being able to give reliable similarity prediction about the samples from classes that were not included in the model training. Koch \etal \cite{koch2015siamese} introduced a deep convolutional model architecture consisting of two convolutional branches sharing weights and joined by a merge layer with $L_1$ cost function describing distance between the two inputs $x_1,x_2$ :
\begin{equation}
\label{equation:l1}
	L_1(x_1,x_2) = |f(x_1) - f(x_2)|
\end{equation}

where $f$ denotes the encoding function. This is combined with a sigmoid activation of the single neuron in the last layer, which maps the output to the range of [0, 1]. The applications of Siamese networks include many areas, most importantly one-shot image recognition, with good benchmark performances achieved on well-known datasets such as Omniglot (written characters recognition for multiple alphabets, 95\% accuracy) and ImageNet (natural image recognition with 20000 classes, 87.8\% accuracy) \cite{VinyalsSiameseNIPS2016}, but also object co-segmentation \cite{SiameseObjectCoSegmentation2018}, object tracking in video scenes \cite{bertinetto2016fullysiamesetracking}, signature verification \cite{BromleySiamense1993}, and even matching resumes to job offers \cite{MaheshwarySiameseResumes2018}. 

\subsection{Data-driven image descriptors}
Several approaches to learning feature descriptors for image matching have been explored, mostly in the field of visual geometry and mapping for image stitching, orientation detection, and similar general-purpose approaches.

Simo-Serra \etal \cite{DeepDescSimoSerra2015} presented a novel point descriptor, whose discriminative feature descriptors are learnt from the real-world, large datasets of corresponding and non-corresponding image patches from the MVS dataset, containing image patches sampled from 3D reconstructions of the Statue of Liberty, Notre Dame cathedral, and Half Dome in Yosemite. The approach is reported to outperform SIFT, while being able to serve as a drop-in replacement for it. The method employs a Siamese architecture of two coupled CNNs with three convolutional layers each, whose outputs are patch descriptors, and an $L_2$ norm of the output difference is minimized between positive patches and maximized otherwise. A similar approach was demonstrated by Zagoruyko and Komodakis \cite{ZagoruykoK15SiamesePatches}, who train a similarity function for comparing image patches directly from the data employing several methods, one of them being a Siamese model with two CNN branches sharing weights, connected at the top by two fully connected layers. 

Yi \etal introduced a method that is intended to serve as a full SIFT replacement, not only as a drop-in descriptor replacement \cite{LIFT}. The deep-learnt approach consists of a full pipeline with keypoint detection, orientation estimation, and feature description, trained in a form of a Siamese quadruple network with two positive (corresponding) input patches, one negative (non-corresponding) patch, and one patch without any keypoints in it. Hard mining of difficult keypoint pairs is employed, similarly to \cite{DeepDescSimoSerra2015}. DeTone \etal \cite{SuperPointDeTone2018} proposed a so-called SuperPoint network and a framework for self-supervised training of interest point detectors and descriptors that are able to operate on the full image as an input, instead of image patches. The method is able to compute both interest points and their descriptors in a single network pass.

Moving to the field of iris recognition, Czajka \etal \cite{CzajkaBSIFIrisDesc2019} have recently employed human-inspired, iris-specific binarized statistical image features (BSIF) from iris image patches derived from an eye-tracking experiment, during which human iris examiners were asked to classify iris image pairs. Data-driven BSIF filters were also studied by Bartuzi \etal for the purpose of person recognition based on thermal hand representations \cite{EwelinkaHandIWBF2018}.

Current literature seems not to offer any iris-driven filters, which would be designed specifically to address post-mortem iris recognition.

\section{Proposed methodology}
\label{sec:methodology}

\subsection{Iris recognition benchmarks}
\label{sec:toolkit}

Previous works assessing post-mortem iris recognition accuracy \cite{TrokielewiczPostMortemICB2016, TrokielewiczPostMortemBTAS2016, TrokielewiczTIFS2018} have employed several iris recognition matchers, including the open-source OSIRIS \cite{OSIRIS}, and three commercially available products: VeriEye \cite{VeriEye}, MIRLIN\footnote{discontinued by the manufacturer} \cite{MIRLIN}, and IriCore \cite{IriCore}. 

In this paper we employ a popular open-source implementation of Daugman's method (OSIRIS), as well as the IriCore commercial product, which was shown to offer the best performance when confronted with post-mortem, heavily degraded iris samples in previous papers \cite{TrokielewiczTIFS2018}. The reasoning behind this choice of iris matchers is to be able to compare the proposed approach against the method that was to date the best performing one in the post-mortem iris recognition setup.

\begin{figure}[t]
\centering
\includegraphics[width=1.02\linewidth]{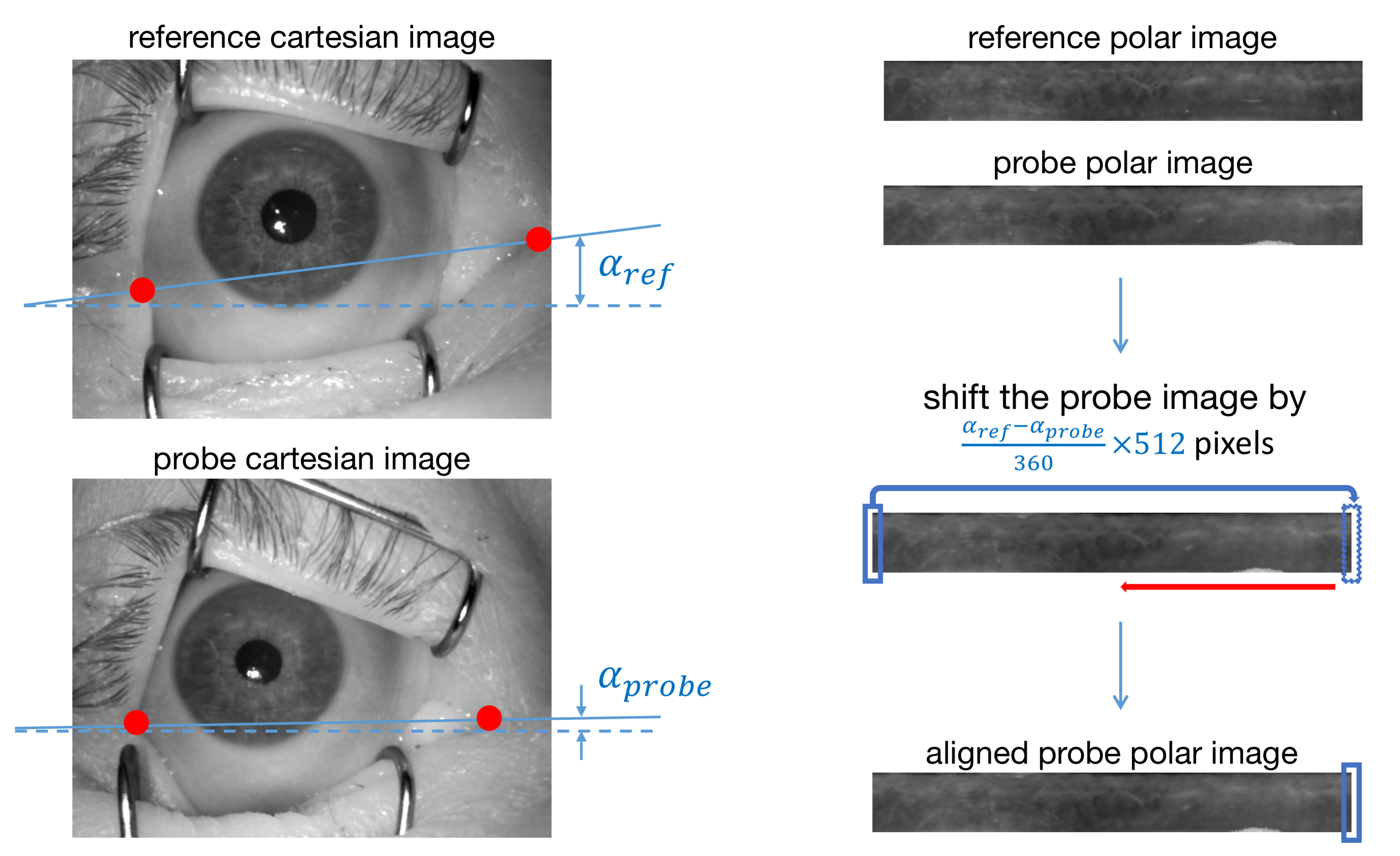}
\caption{Aligning the polar-coordinate polar iris images based on eye corner location to compensate rotation of the camera during image acquisition.}
\label{fig:code_shifting}
\end{figure}

\begin{figure}[t]
\centering
\includegraphics[width=1.02\linewidth]{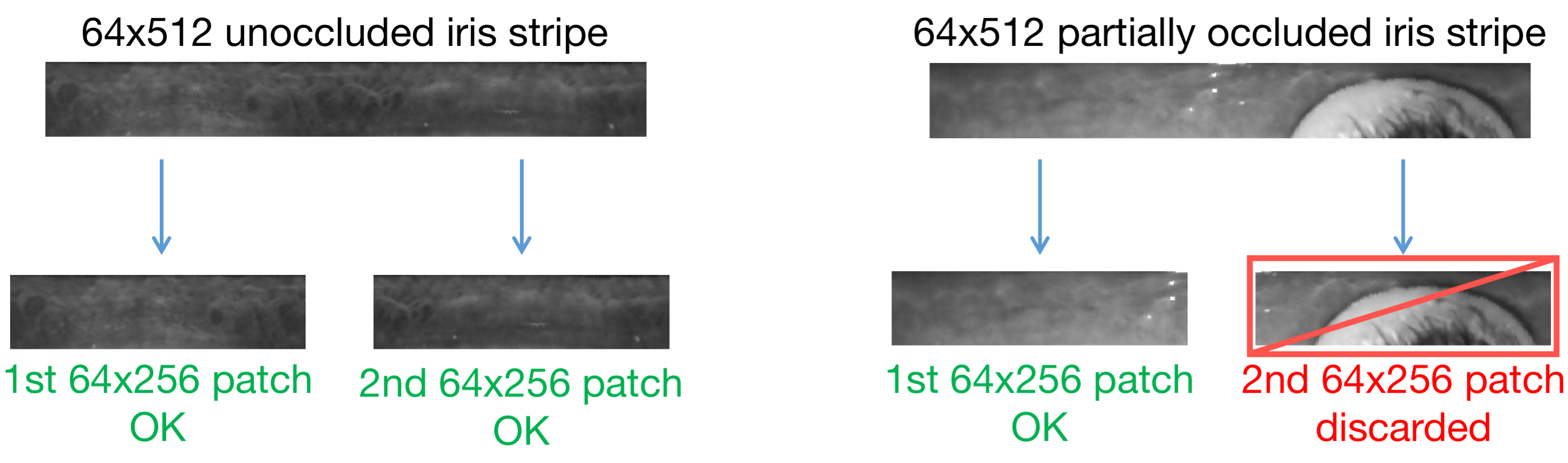}
\caption{Scheme for iris patch curation.}
\label{fig:patch_creation}
\end{figure}

\subsection{Image segmentation}
\label{sec:segmentation}
\begin{figure}[t!]
\centering
	\begin{subfigure}[t]{0.23\textwidth}
		\includegraphics[width=\textwidth]{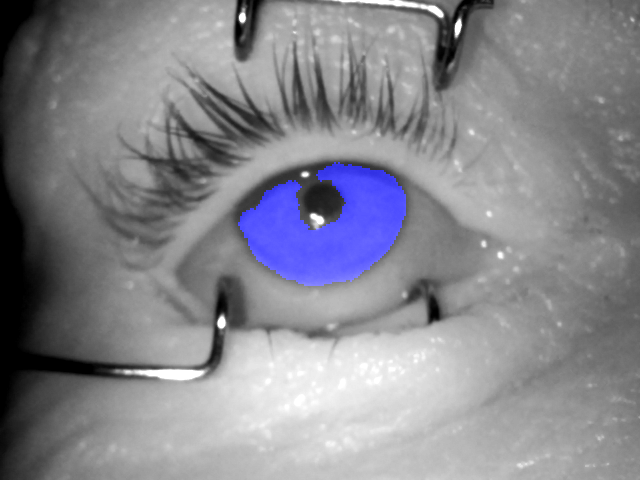}
		\caption{Example segmentation result for the original model \cite{TrokielewiczIMAVIS2019arxiv}.}
	\end{subfigure}\hfill
		\begin{subfigure}[t]{0.23\textwidth}
		\includegraphics[width=\textwidth]{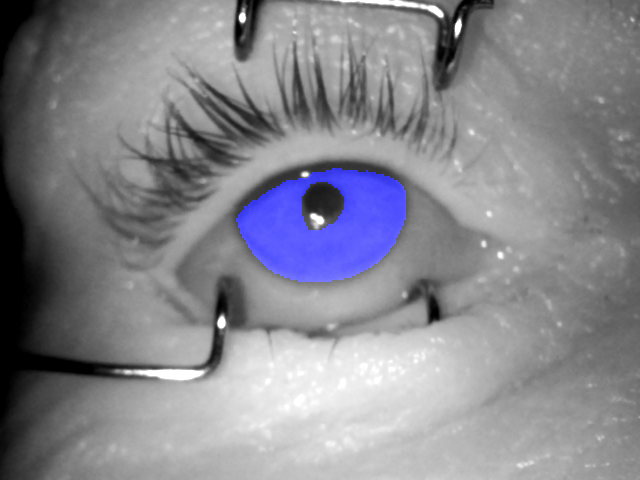}
		\caption{The same sample processed with the new segmentation model.}
	\end{subfigure}	
\caption{Segmentation results obtained for an example iris image using the model from earlier works and from the new, refined model.}
\label{fig:newSegmentationExample}
\end{figure}

It has been shown that a significant fraction of post-mortem iris recognition errors can be attributed to the badly executed segmentation stage \cite{TrokielewiczPostMortemBTAS2016, TrokielewiczTIFS2018}. Trokielewicz \etal \cite{TrokielewiczIMAVIS2019arxiv} proposed an open-source CNN-based semantic segmentation model for predicting binary masks of post-mortem iris images (trained with iris images obtained from cadavers, elderly people, as well as ophthalmology patients), which in this paper has been further fine-tuned with more diverse datasets, including ISO-compliant images, as well as low quality, visible-light iris samples. In addition to the databases used for training the previous model, we have also employed several iris datasets with their corresponding ground truth masks, including the Biosec baseline corpus \cite{biosec} (1200 images), the BATH database\footnote{\scriptsize\url{http://www.bath.ac.uk/elec-eng/research/sipg/irisweb/}} \cite{bath} (148 images), the ND0405 database\footnote{\label{cvrl}\scriptsize\url{https://cvrl.nd.edu/projects/data/}} (1283 images), the UBIRIS database \cite{ubiris} (1923 images), and the CASIA-V4-Iris-Interval database\footnote{\scriptsize\url{http://www.cbsr.ia.ac.cn/english/IrisDatabase.asp}} (2639 images). This allowed to obtain more coherent, smoother predictions, as shown in Fig. \ref{fig:newSegmentationExample}. The model was trained with the SGDM optimizer with momentum = 0.9, learning rate = 0.001, $L_2$ regularization = 0.0005, for 120 epochs with batch size of 4. The trained model along with sample codes is made available with this paper\footnote{\scriptsize\url{link to the repo temporarily removed to make this submission anonymous}}.

\subsection{Training data}
To train our new, post-mortem-aware iris feature descriptor, we use NIR iris images from publicly available {\it Warsaw-BioBase-Postmortem-Iris-v1.1}  and {\it Warsaw-BioBase-Postmortem-Iris-v2} databases, which were processed with the new segmentation model (cf. Sec. \ref{sec:segmentation}) and normalized to come up with polar iris images $512\times64$ pixels in size. Normalization stage included Hough Transform-based circle fitting to approximate inner and outer iris boundaries, and all pixels annotated as non-iris texture pixels found by segmentation model inside the iris annulus were discarded from feature extraction.

\subsection{Evaluation data}
\label{sec:Database}
For evaluation of the proposed method an in-house, newly collected database of post-mortem iris images is used. It comprises post-mortem images taken from 40 subjects with a total of 1094 near-infrared images and 785 visible-light images, collected up to 369 hours since demise. This dataset is subject-disjoint to the data used both in the training of the segmentation CNN model, and it was collected following the same acquisition protocol as in collecting the Warsaw data.

\begin{figure*}[t]
\centering
\includegraphics[width=0.78\linewidth]{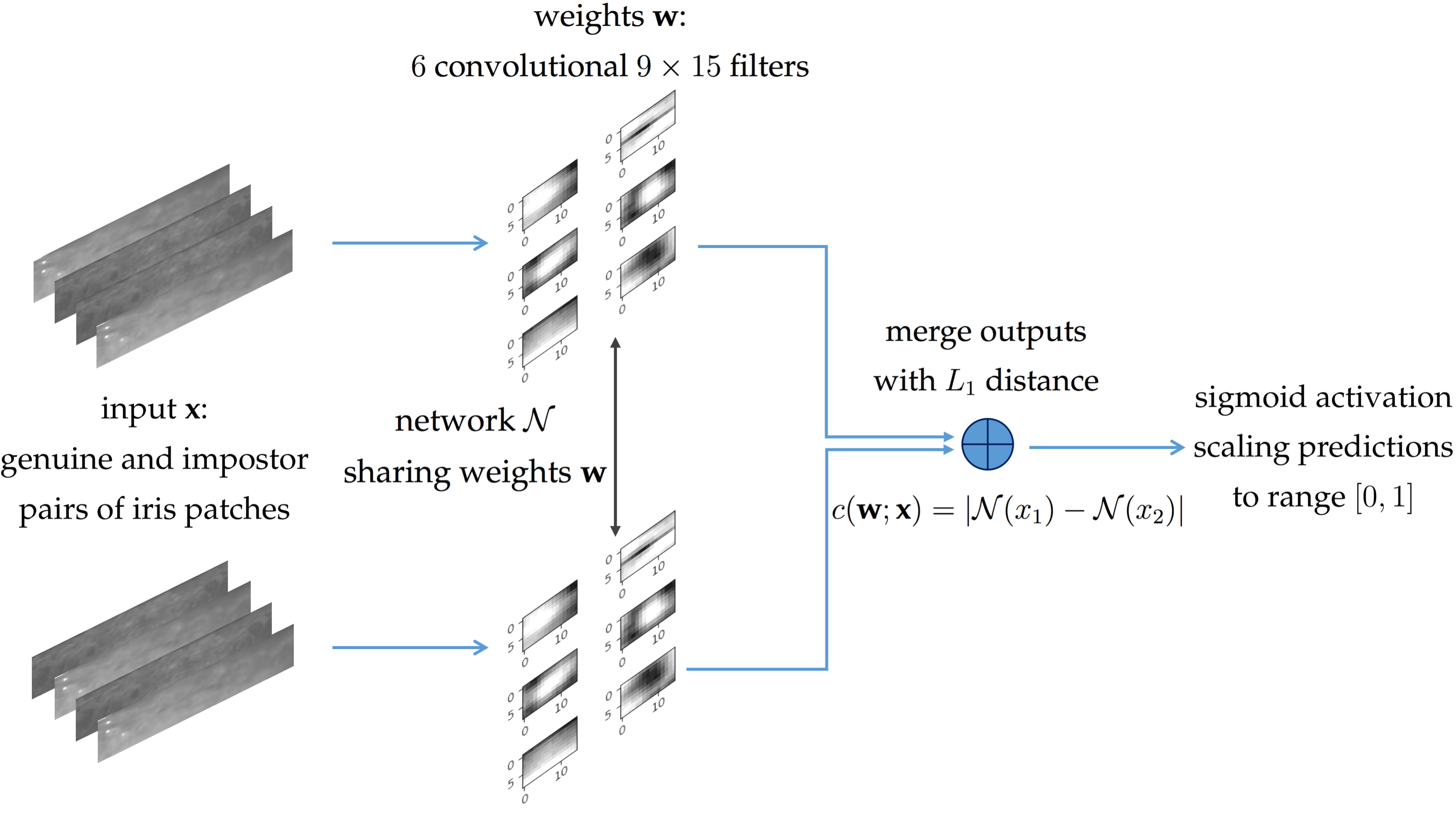}
\caption{Shallow Siamese network used for learning the iris-specific feature representation.}
\label{fig:siamese_net}
\end{figure*}

\subsection{Pre-processing of the training data}
\label{sec:preprocessing}

The first step in preparing the training data is to ensure spatial correspondence between iris features in multiple images of the same eye. Note that post-mortem iris images can have arbitrary orientation, as the cadavers may be approached by the operator taking the scans from different directions, for instance, depending on the crime scene layout. Leaving these pictures unaligned would end up with forcing the network to learn inappropriate kernels. This is done by doing within-class alignment of normalized iris images by canceling the mutual horizontal shifts between images that reflects eyeball rotation in the Cartesian coordinate system, Fig. \ref{fig:code_shifting}. We performed here the image alignment procedure, which involved manual annotations of the eye corners. This allowed to calculate a relative rotation of the eyeball represented in the two images, and in turn the amount of pixels, by which the polar image must be shifted. Iris recognition methods typically shift iris codes in the matching stage to compensate for eyeball rotation, instead of shifting images. Therefore there is no justification to make the neural network learn how to discriminate between irises that are not ideally spatially aligned.

The resulting aligned polar iris images were then subject to examination in respect to the amount of occlusions caused by eyelids or eyelashes. This is done to ensure that the network will use iris-specific and not occlusion-specific regions in development of kernels. Thus, to ensure good quality of the training data, we divide each polar iris image into two \emph{patches}. Since in our data eyelid occlusions are only present either on the left or on the right portion of the polar iris image (this is specific to post-mortem data acquisition protocol in our collection), this enables discarding such samples while at the same time saving the other, unaffected patch, Fig. \ref{fig:patch_creation}. A total of 1801 patches were extracted for training.  


\subsection{Model architecture and filter learning}
\label{sec:architecture}

\begin{figure}[t!]
\centering
	\begin{subfigure}[t]{0.2\textwidth}
		\includegraphics[width=\textwidth]{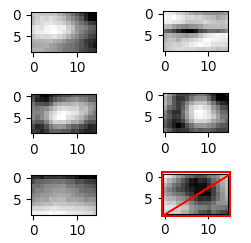}
		\caption{Filter kernels}
	\end{subfigure}\hskip1mm
		\begin{subfigure}[t]{0.25\textwidth}
		\includegraphics[width=\textwidth]{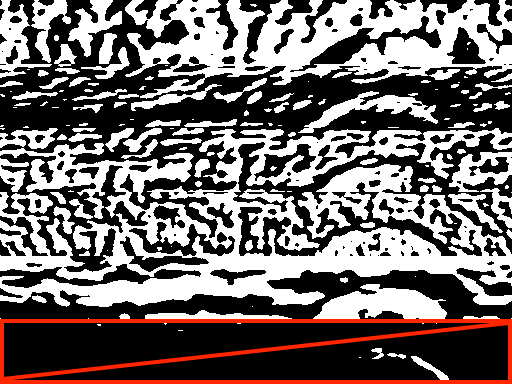}
		\caption{Example iris codes}
	\end{subfigure}\\\vskip1mm
	\begin{subfigure}[t]{0.52\textwidth}
		\includegraphics[width=\linewidth]{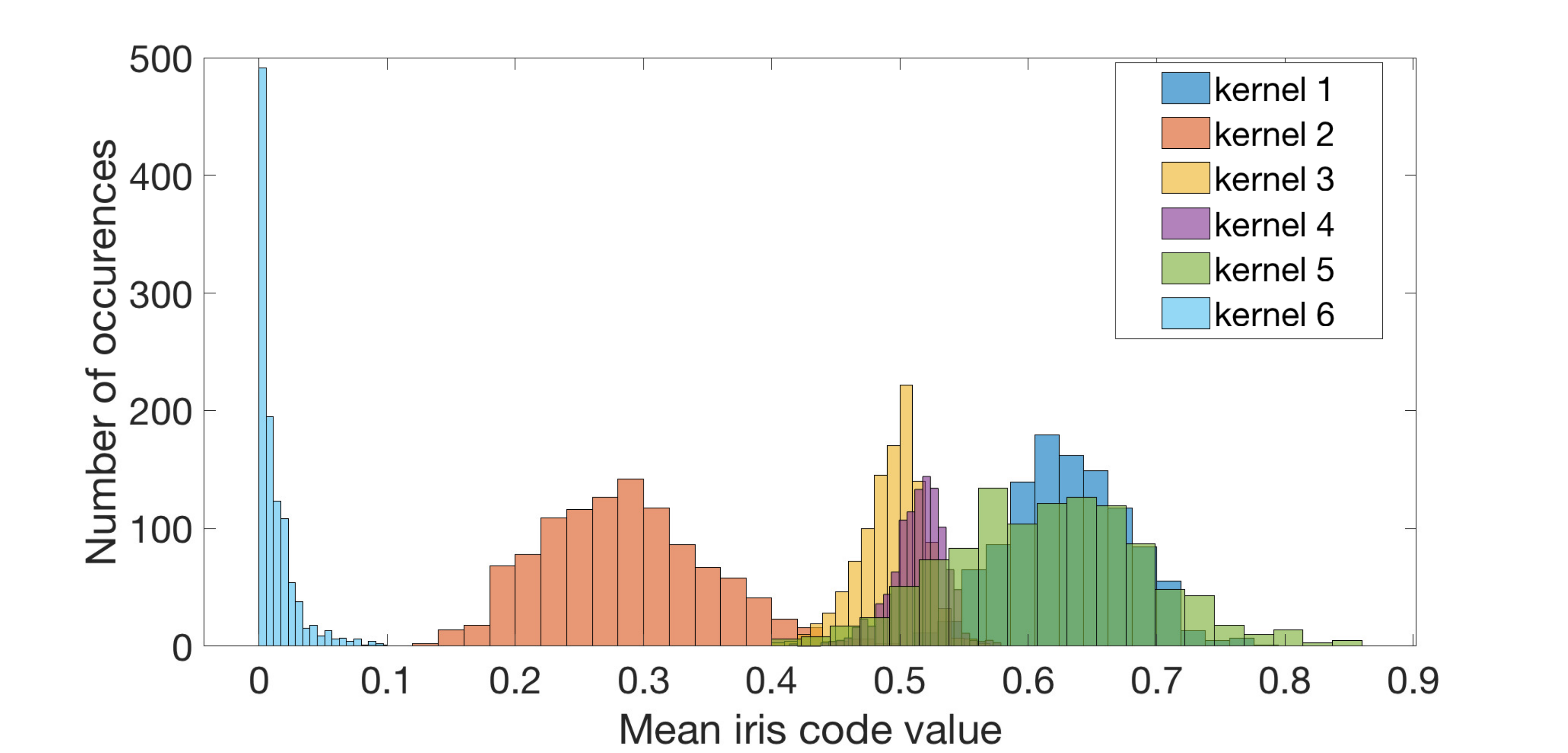}
		\caption{Mean iris code value distributions}
	\end{subfigure}
\caption{Learnt kernels from the Siamese network (a), an example set of iris codes they produce (b), and distributions of mean iris code values across the data set (c). The kernel no. 6 is discarded as it produces iris codes with almost all elements equal to 0.}
\label{fig:siamese_filtersAndCodes}
\end{figure}

For learning the iris-specific filters we use a shallow Siamese architecture composed of two branches, each responsible for encoding one image from the image pair being compared, comprising a single convolutional layer with 6 kernels of size $9\times15$ (the size of the smallest OSIRIS filter), as shown in Fig. \ref{fig:siamese_net}. In our experiments, the $9\times15$ OSIRIS kernels resulted in better post-mortem iris recognition performance than either $9\times27$ or $9\times51$ (also found in OSIRIS). The weights are shared between the two branches. Following the convolutional layers is a \emph{merge} layer calculating the $L_1$ distance (\ref{equation:l1}) between the two sets of features from the convolutional layer. A single neuron with a sigmoid activation function is then applied to yield a prediction from the range [0, 1], with 0 being a perfect match between the two images, and 1 being a perfect non-match.

The training data is passed to the network in batches containing 32 pairs of iris patches, out of which 16 are genuine, and 16 are impostor pairs, randomly sampled without replacement from the dataset during each training iteration, with a total of 20000 iterations. ADAM optimizer \cite{ADAM} is used with the learning rate $lr=0.0006$ to minimize the loss function.

\subsection{Filter set optimization}

The learnt filter kernels, together with example iris codes that they produce, as well as distributions of mean iris code values produced by each of them are illustrated in Fig. \ref{fig:siamese_filtersAndCodes}. By analyzing the distributions of mean iris code values obtained by each of the new filters, we see that codes produced by the sixth filter do not represent the iris well, as most of the texture information is lost during encoding, resulting in a mostly zeroed iris code. This filter is discarded from all further experiments.
 
Notably, employing only iris-specific filter kernels instead of those found in OSIRIS did not yield better results -- perhaps due to the fact that regular iris texture is well represented using the conventional Gabor wavelets, and the newly learnt filters are necessary to boost the performance for difficult, decay-affected samples. To utilize these new filters, and to offer an advantage over the baseline method, a modification of the OSIRIS Gabor filter bank was performed to propose a {\bf hybrid filter bank} comprising a combination of Gabor wavelets and the post-mortem-iris-specific kernels.

The filter selection was solved by using Sequential Feature Selection, with a combination of Sequential Forward Selection (SFS) and Sequential Backward Selection (SBS), which involve adding the most discriminant features to the classifier, while removing the least discriminatory ones. During the feature selection procedure, post-mortem-specific filters were added to the original OSIRIS filter bank, whereas those OSIRIS filters, which do not contribute to decreasing the error rates were removed. The error metric minimized during the feature selection is the EER obtained for samples acquired up to 60 hours post-mortem.

The feature selection procedure can be enclosed in the following steps, starting with the original, unmodified OSIRIS filter bank comprising six Gabor wavelets:
\begin{itemize}
	\item [] {\bf Step 1.} Calculate the performance obtained using the current filter bank and each of the Siamese filters added independently.
	\item [] {\bf Step 2.} \textcolor{mygreen}{\bf Perform SFS by adding the most contributing filter to the filter bank.}
	\item [] {\bf Step 3.} Calculate the performance obtained using the filter bank obtained in the previous step with each of the OSIRIS filters removed independently.
	\item [] {\bf Step 4.} \textcolor{myred}{\bf Perform SBS by removing the least contributing filter from the filter bank.}
	\item [] {\bf Step 5.} $\rightarrow$ Go back to Step 1 and repeat until the error metric stops improving.
\end{itemize}

\begin{figure*}[t]
\centering
\includegraphics[width=0.66\textwidth]{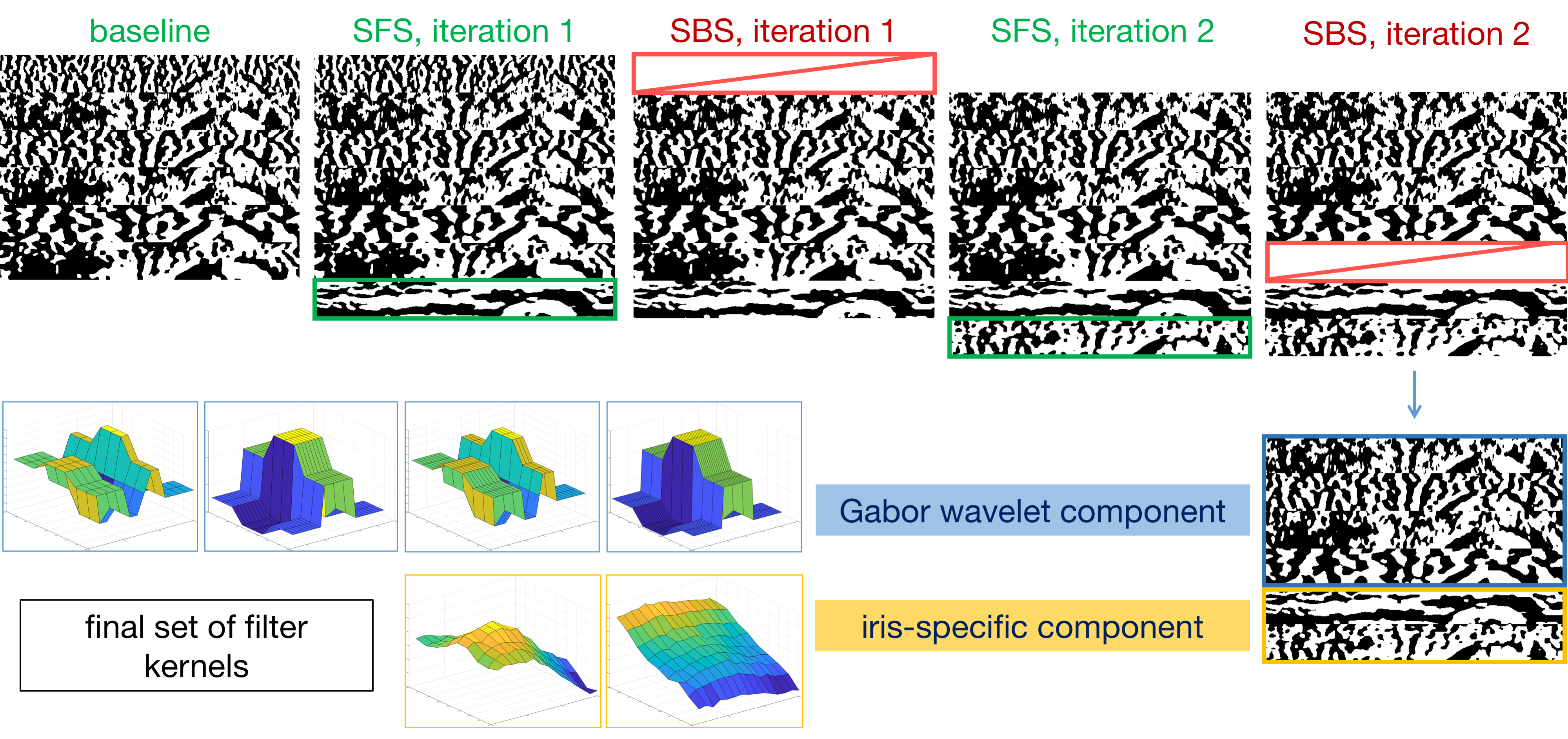}
\caption{Filter selection for the new filter bank using Sequential Forward Selection and Sequential Backward Selection. Iris codes for an example iris produced by the proposed hybrid filter bank at different stages of filter selection are also shown.}
\label{fig:siamese_featureSelection}
\end{figure*}

\begin{figure}[t]
\centering
\includegraphics[width=\linewidth]{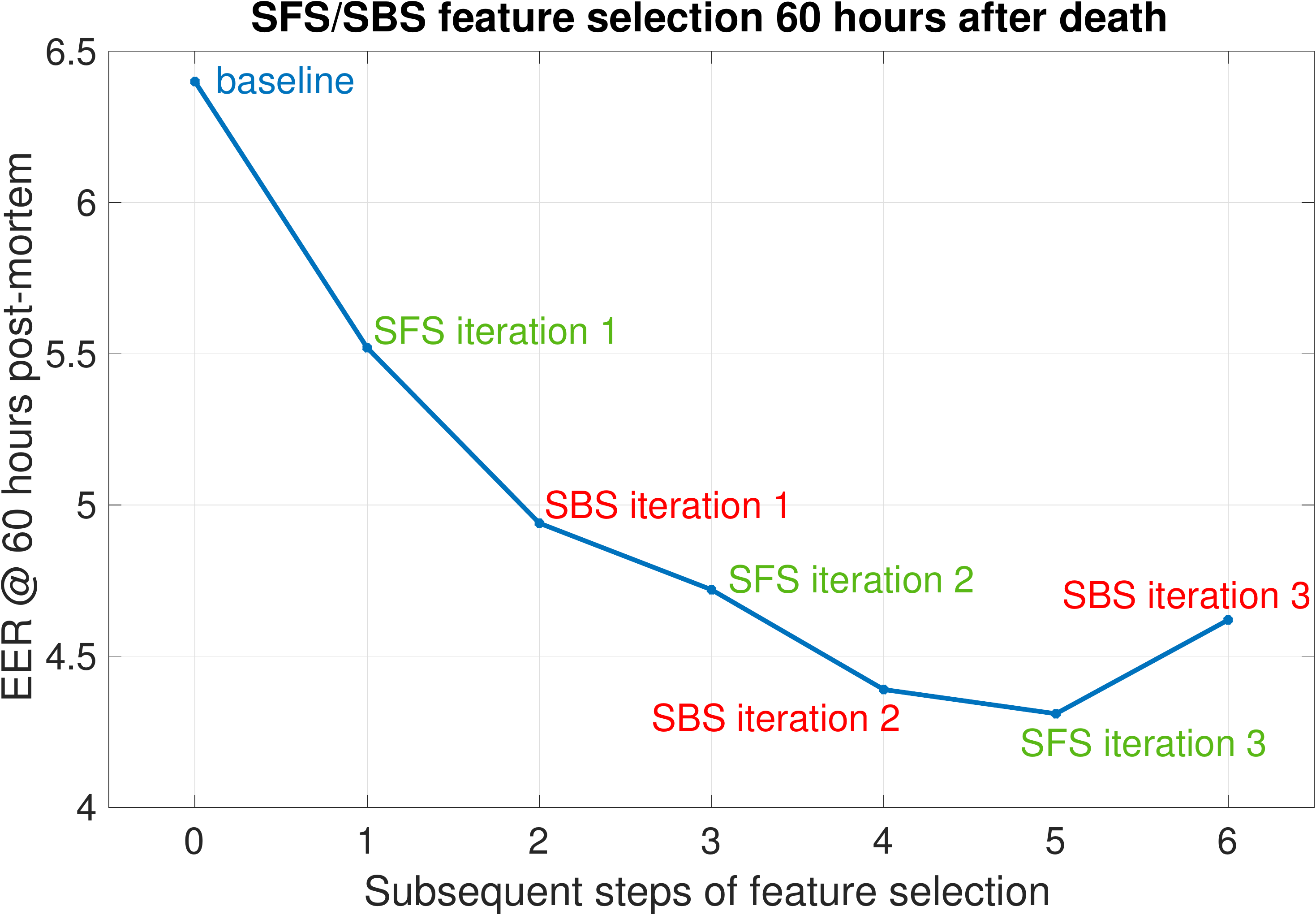}
\caption{SFS/SBS filter selection: Equal Error Rates in subsequent iterations of the procedure are shown. The filter selection is stopped at SBS iteration no. 2.}
\label{fig:SFS}
\end{figure}

After the SFS/SBS feature selection procedure involving four iterations of SFS and SBS, Fig. \ref{fig:siamese_featureSelection}, the EER was decreased by almost a third, from 6.40\% obtained for the 60 hours post-mortem time horizon for the original OSIRIS filter bank, to 4.39\% obtained for the new, hybrid filter bank, Fig. \ref{fig:SFS}. This filter bank is then used in the final testing of our iris recognition pipeline with the new, subject-disjoint data. The final set of filter kernels is shown in Fig. \ref{fig:siamese_featureSelection}.


\section{Results and discussion}

\subsection{Testing data and methodology}
During testing, we generate all possible genuine and impostor scores between images that were captured up to a certain time horizon after demise. Nine different time horizons are considered, namely: up to 12, 24, 48, 60, 72, 110, 160, 210, and finally up to 369 hours post-mortem, which encompasses all available testing data. Every one of these 9 experiments is repeated for comparison scores obtained from (a) the original OSIRIS software, (b) the commercial IriCore matcher, (c) the modified OSIRIS including the new image segmentation, and finally (d) the modified OSIRIS, which includes both the new segmentation and the new feature representation stage.  

\subsection{Recognition accuracy}
\begin{figure*}[h!]
\centering
\includegraphics[width=0.33\linewidth]{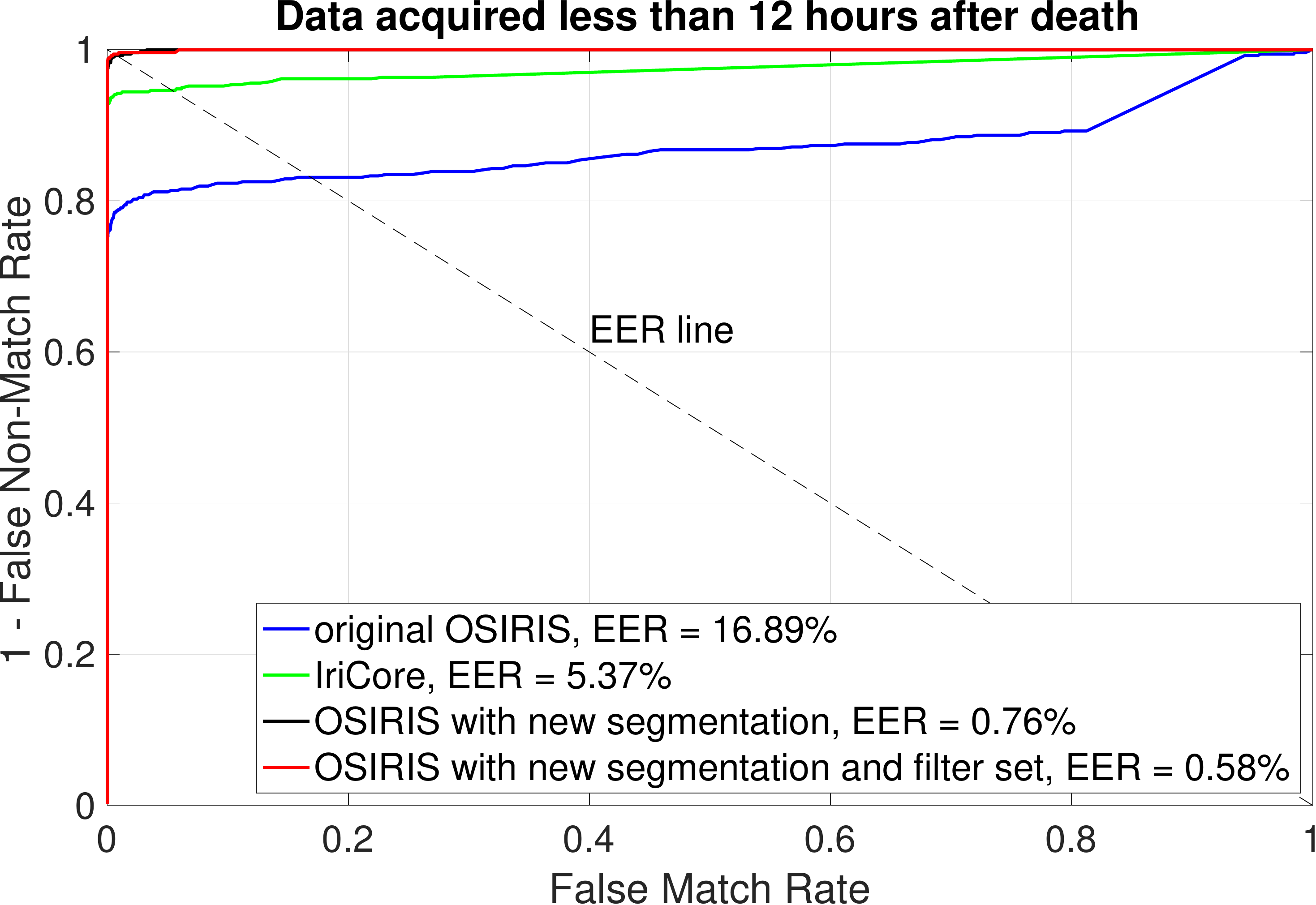}\hfill
\includegraphics[width=0.33\linewidth]{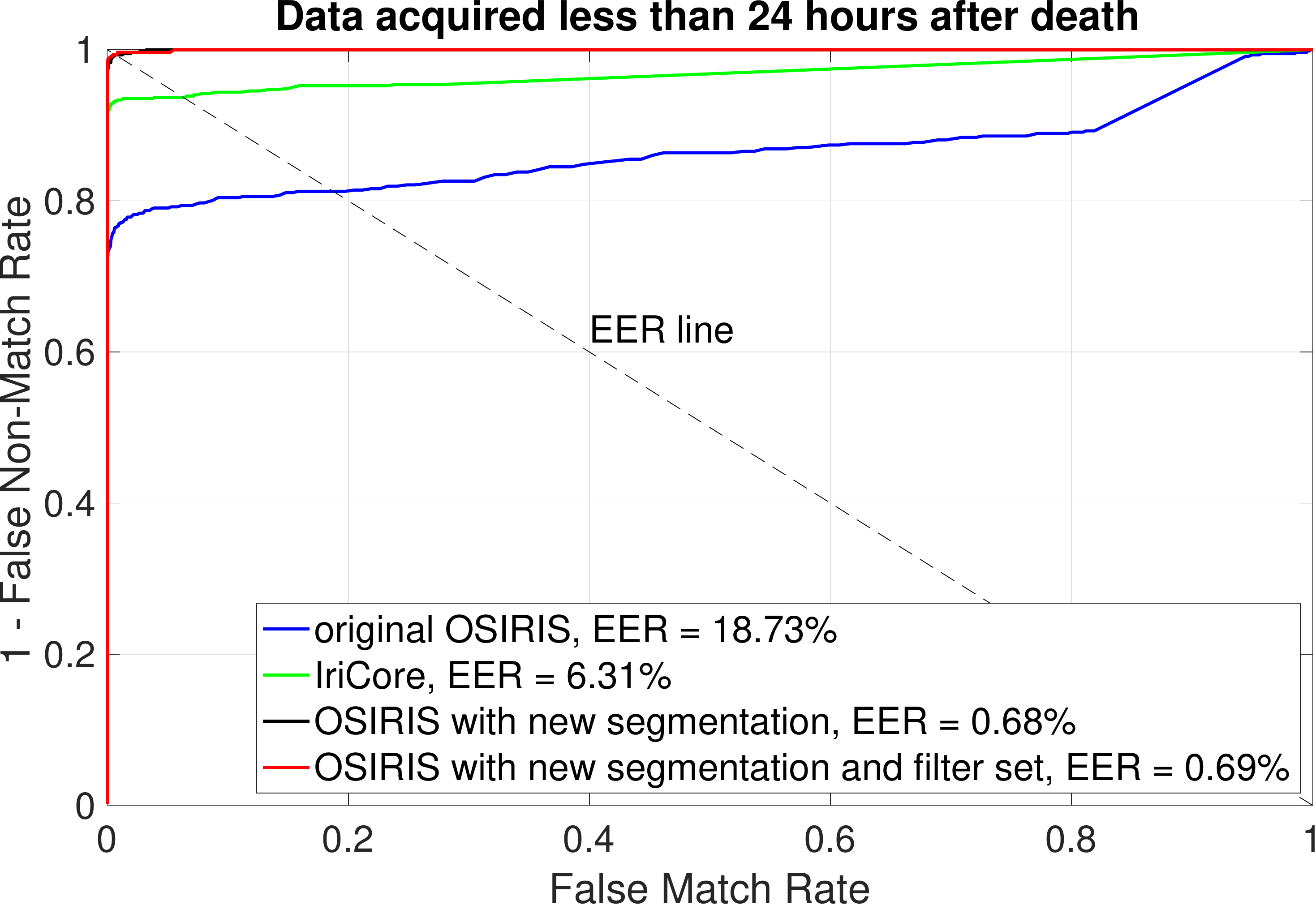}\hfill
\includegraphics[width=0.33\linewidth]{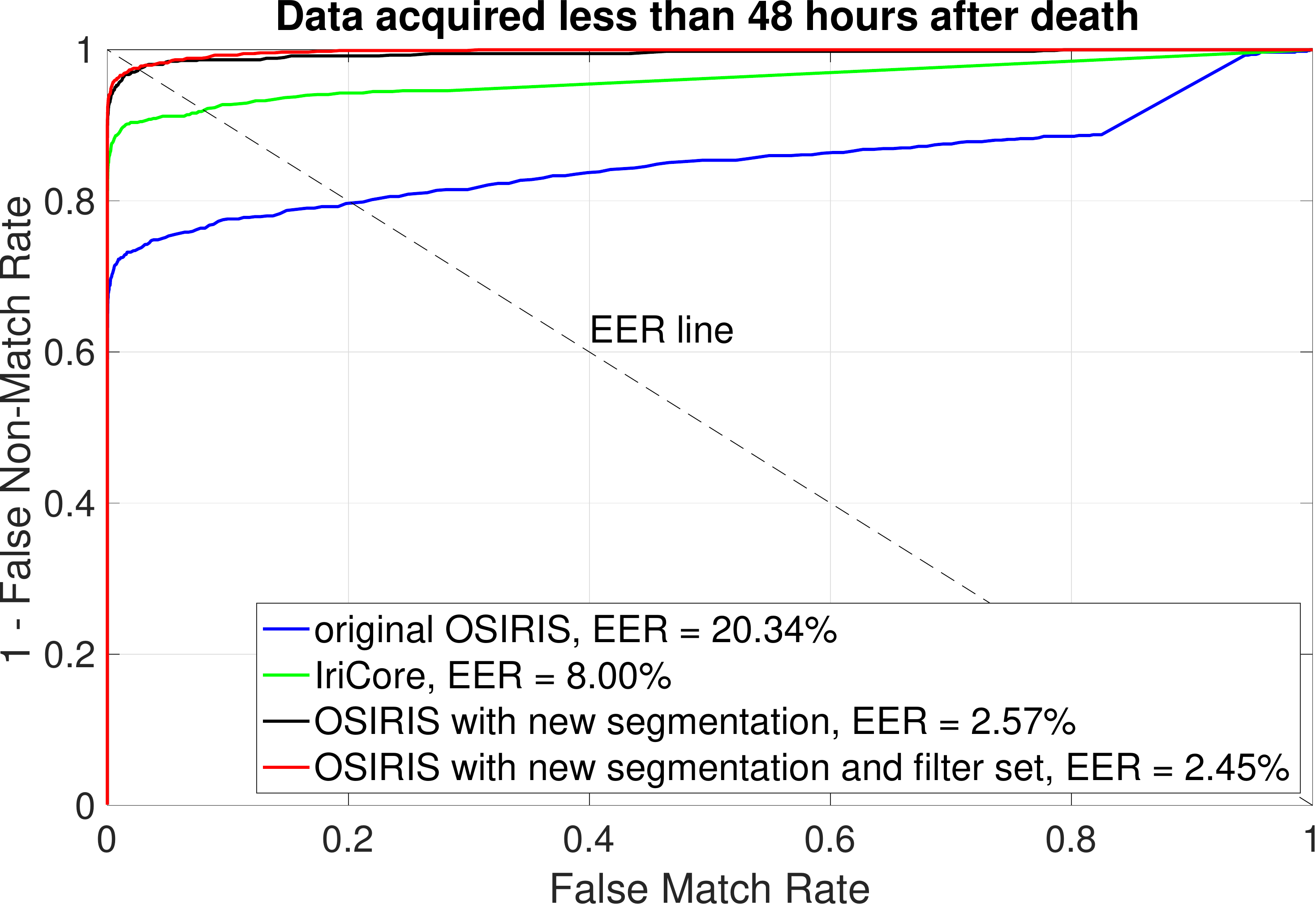}\hfill
\caption{ROC curves obtained when comparing post-mortem samples with different observation time horizons: 12, 24, and 48 hours post-mortem, plotted for two baseline iris recognition methods OSIRIS (blue) and IriCore (green), OSIRIS with CNN-based segmentation (black), as well as OSIRIS with both the improved segmentation and new filter set (red).}
\label{fig:ROCs:cold_short_newfilters}
\end{figure*}

\begin{figure*}[h!]
\centering
\includegraphics[width=0.33\linewidth]{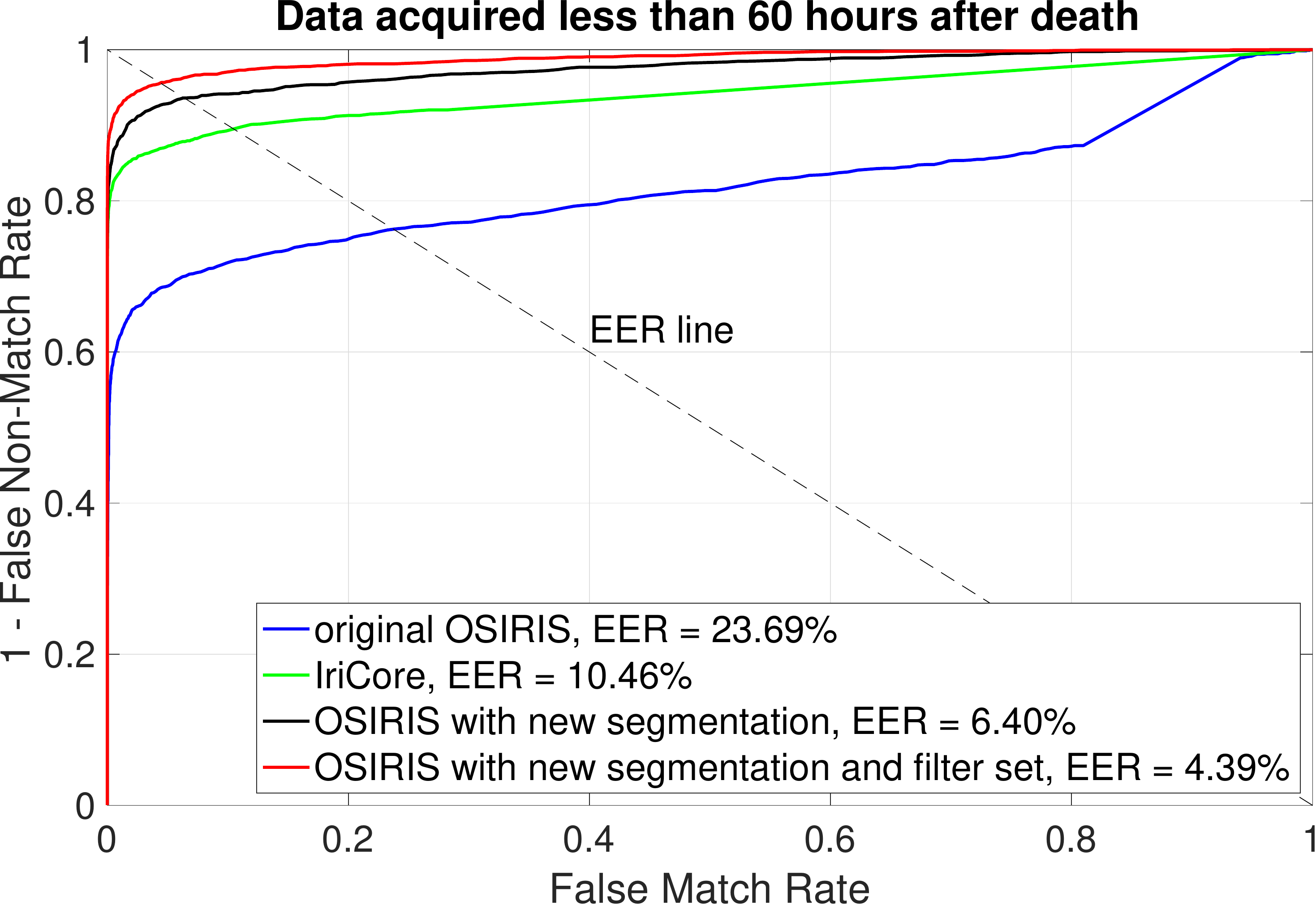}\hfill
\includegraphics[width=0.33\linewidth]{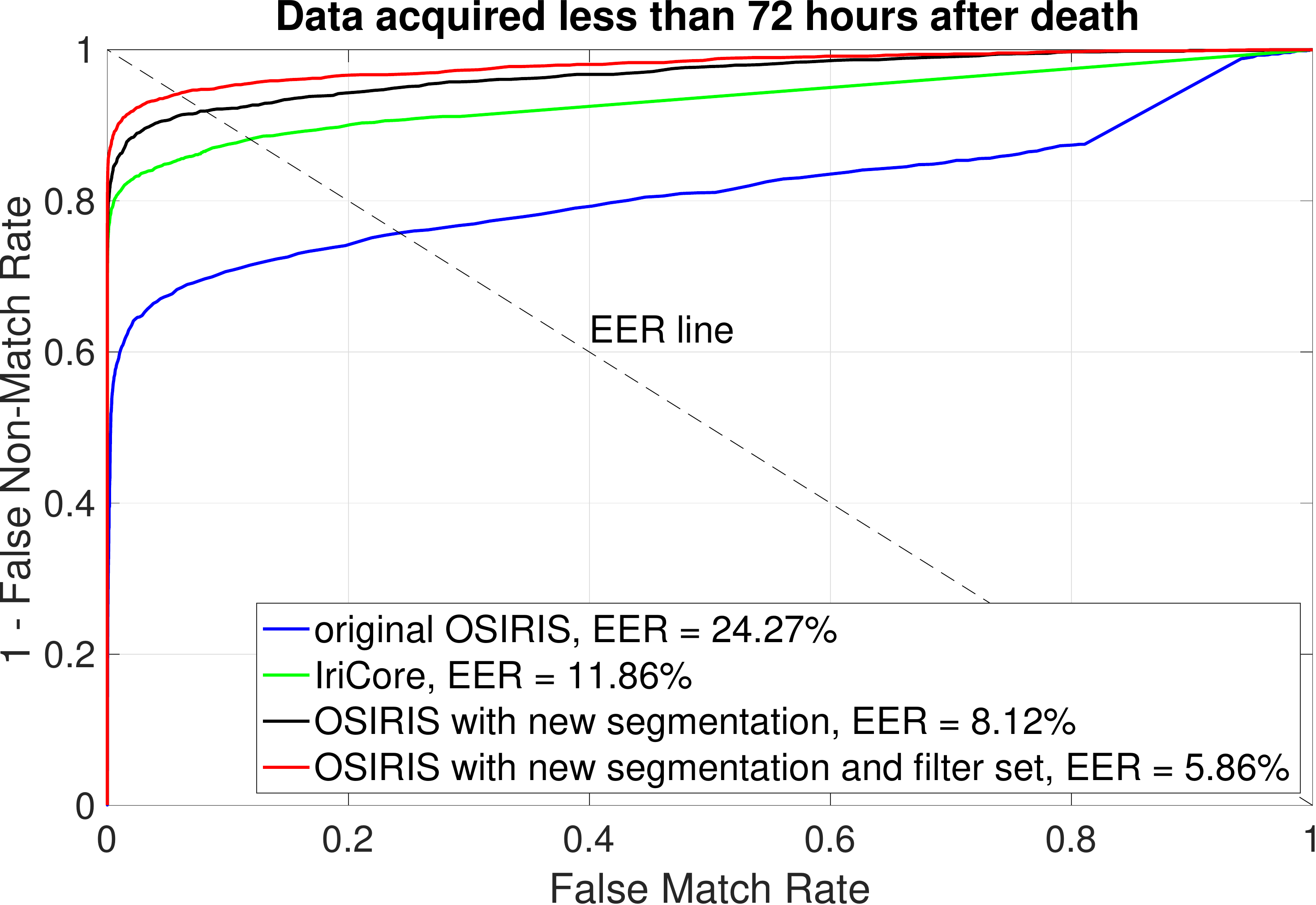}\hfill
\includegraphics[width=0.33\linewidth]{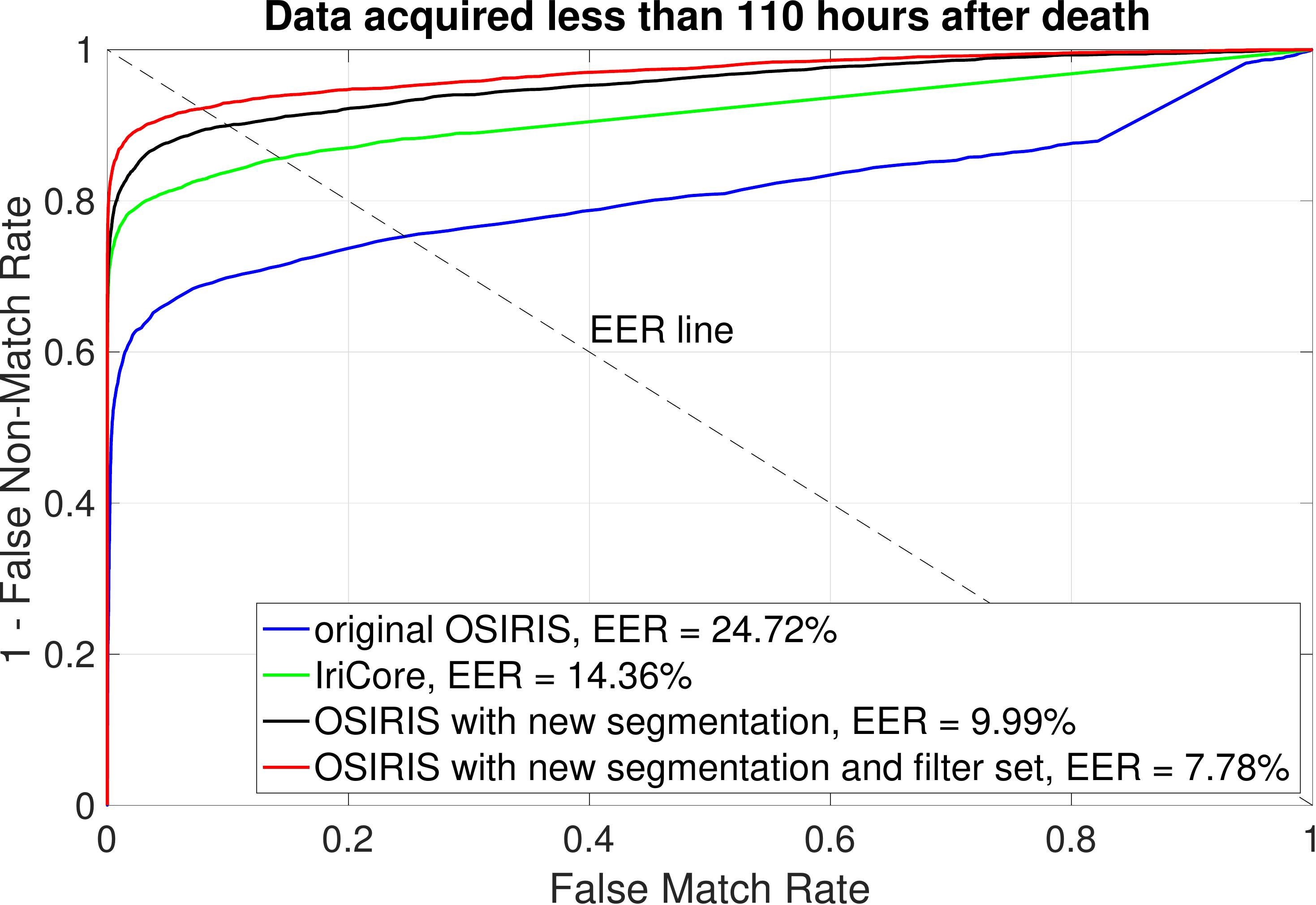}\hfill
\caption{Same as in Fig. \ref{fig:ROCs:cold_short_newfilters}, but for samples collected up to 60, 72, and 110 hours post-mortem.}
\label{fig:ROCs:cold_mid_newfilters}
\end{figure*}

\begin{figure*}[h!]
\centering
\includegraphics[width=0.33\linewidth]{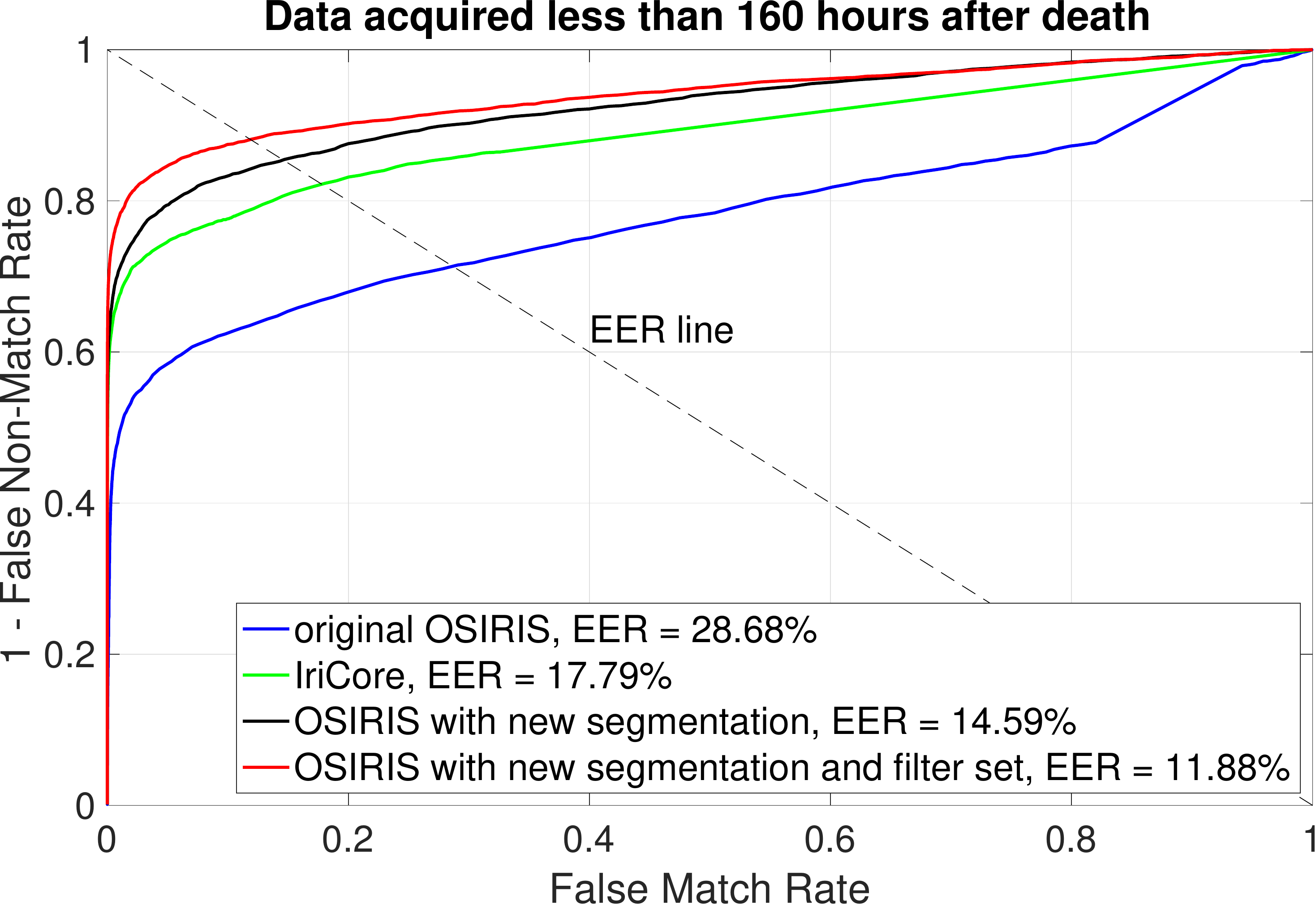}\hfill
\includegraphics[width=0.33\linewidth]{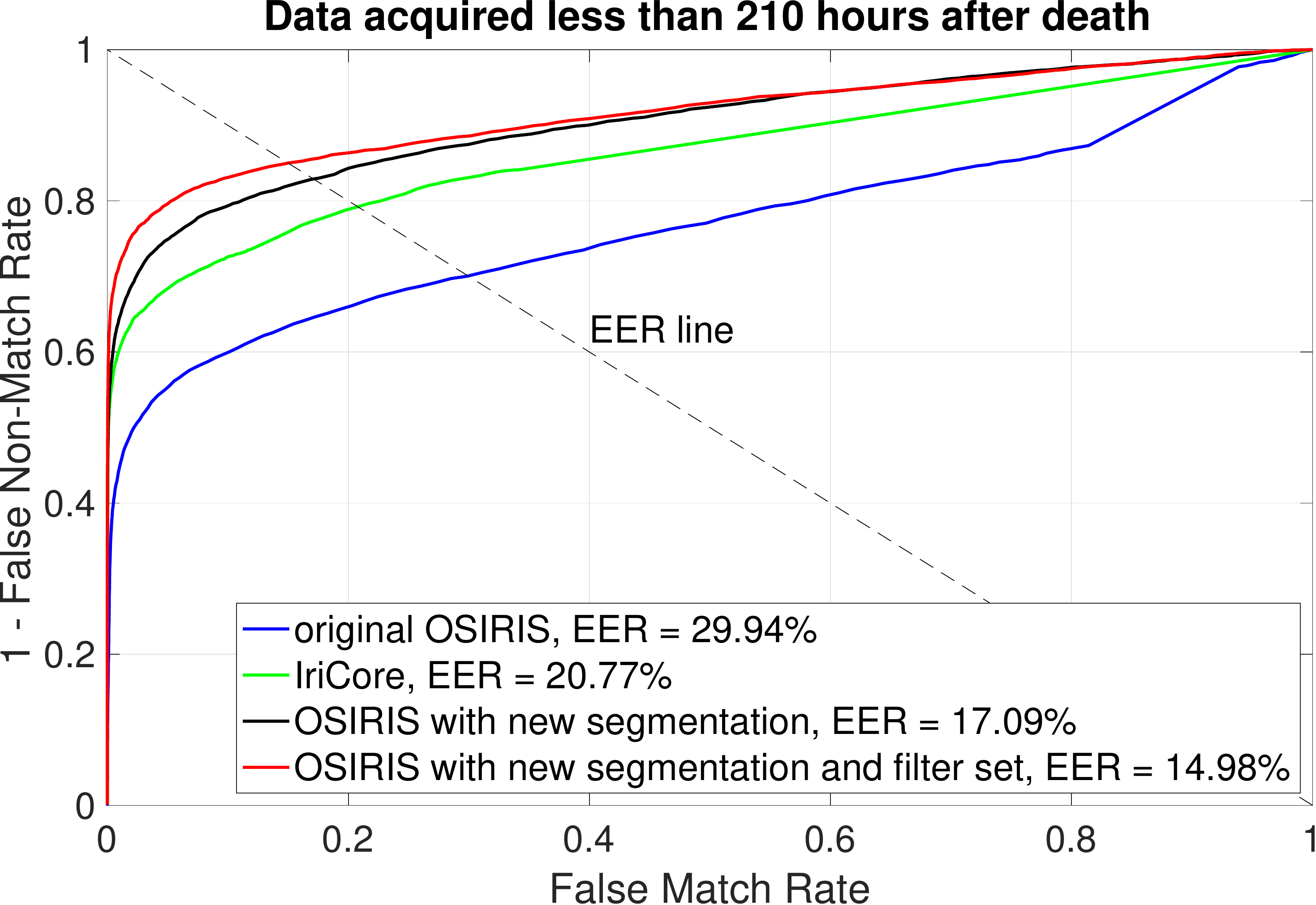}\hfill
\includegraphics[width=0.33\linewidth]{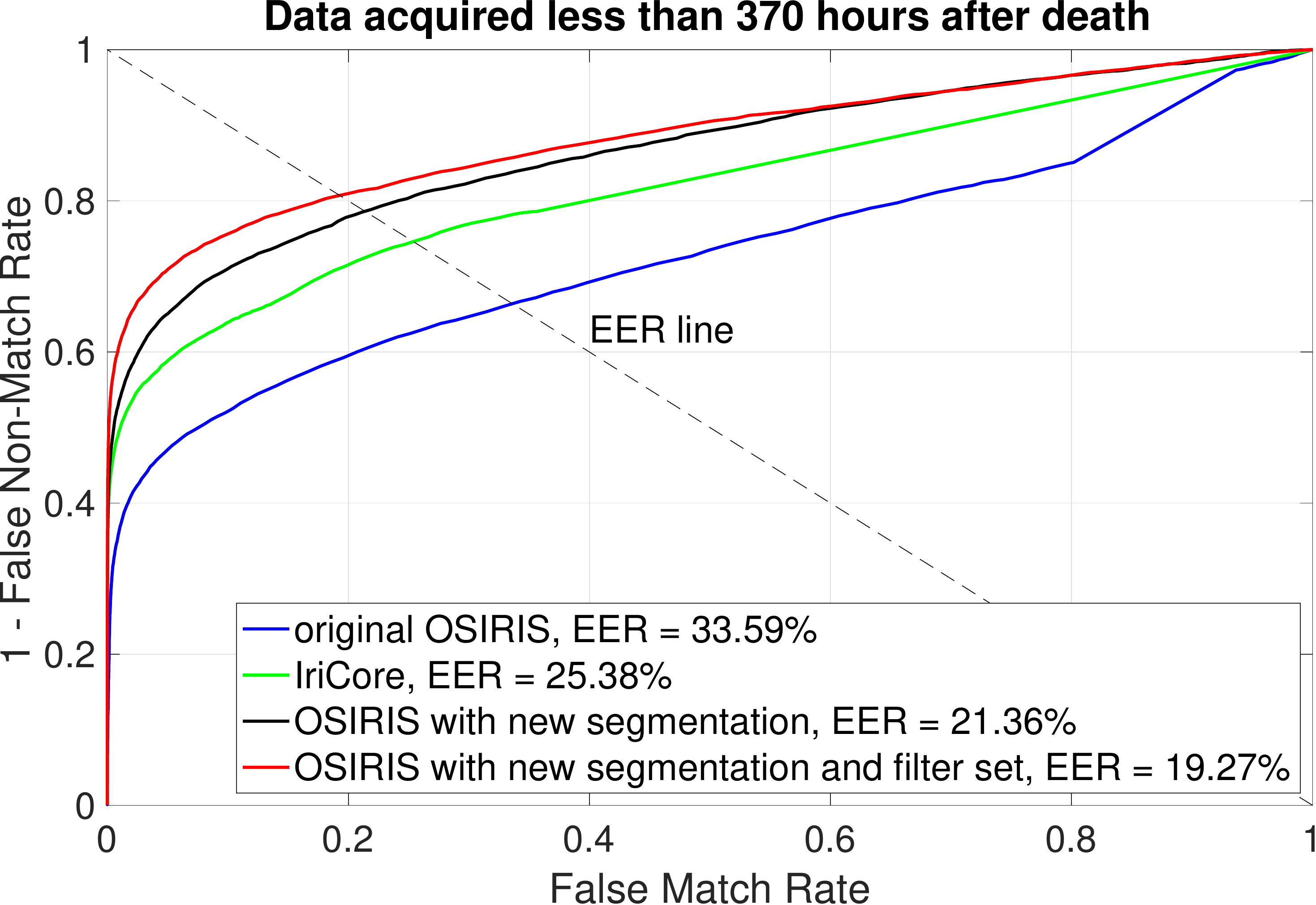}\hfill
\caption{Same as in Fig. \ref{fig:ROCs:cold_short_newfilters}, but for samples collected up to 160, 210, and 369 hours post-mortem.}
\label{fig:ROCs:cold_long_newfilters}
\end{figure*}

Figures \ref{fig:ROCs:cold_short_newfilters}-\ref{fig:ROCs:cold_long_newfilters} present ROC curves obtained using the newly introduced filter bank, compared against ROCs corresponding to the best results obtained when only the segmentation stage is replaced with the proposed modifications.

When analyzing graphs presented in Fig. \ref{fig:ROCs:cold_short_newfilters}, which correspond to samples collected up to 12, 24, and 48 hours post-mortem, we do not see considerable improvements in recognition performance measured by the EER, which, compared against those EERs obtained only with the modified segmentation stage, are EER=0.76\%$\rightarrow$0.58\%, 0.69\%$\rightarrow$0.68\%, and 2.57\%$\rightarrow$2.45\%, respectively. However, the shapes of the red graphs corresponding to the scores obtained with the new filter bank show an improvement over the black graphs in the low FMR registers, meaning that the proposed system offers higher recognition rates in situations, when very few false matches are allowed.

Moving to more distant post-mortem sample capture time horizons, Fig. \ref{fig:ROCs:cold_mid_newfilters}, the advantage of the proposed method becomes clearly visible in both the decreased EER values, as well as in the shapes of the ROC curves. Applying domain-specific filters allowed to reduce EER from 6.40\% to 4.39\%, 8.12\%$\rightarrow$5.86\%, and 9.99\%$\rightarrow$7.78\%, for samples acquired less than 60, 72, and 110 hours post-mortem, respectively. 

Finally, Fig. \ref{fig:ROCs:cold_long_newfilters} presents ROC curves obtained for samples collected during the three longest subject observation time horizons, namely up to 160, 210, and 369 hours after death. Here as well, a visible improvement offered by the new feature representation scheme is reflected in the decreased EER values -- 14.59\%$\rightarrow$11.88\% for samples collected up to 160 hours, 17.09\%$\rightarrow$14.98\% for those captured up to 210 hours, and 21.36\%$\rightarrow$19.27\% for the longest and most difficult set, encompassing images acquired up to 369 hours (more than 15 days).


\subsection{False Non-Match Rate dynamics}
\begin{figure*}[htb]
\centering
\includegraphics[width=0.68\linewidth]{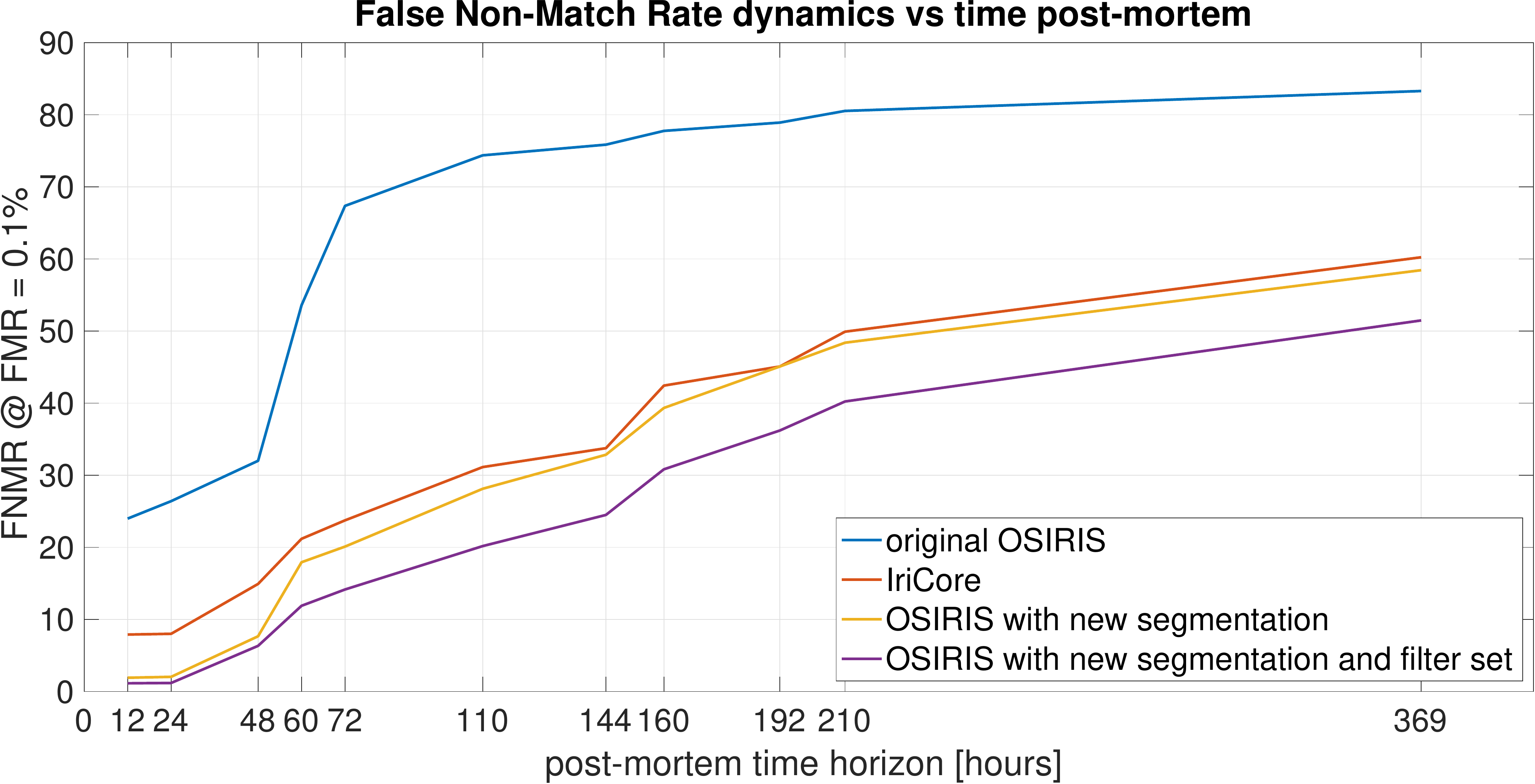}
\caption{False Non-Match Rates (FNMR) as a function of post-mortem sample capture horizon for a set False Match Rate (FMR) of 0.1\%, plotted for OSIRIS (blue), IriCore (red), OSIRIS with new segmentation (yellow), and OSIRIS with both new segmentation and new filter set (violet).}
\label{fig:FNMR01_new}
\end{figure*}

\begin{figure*}[h!]
\centering
\includegraphics[width=0.68\linewidth]{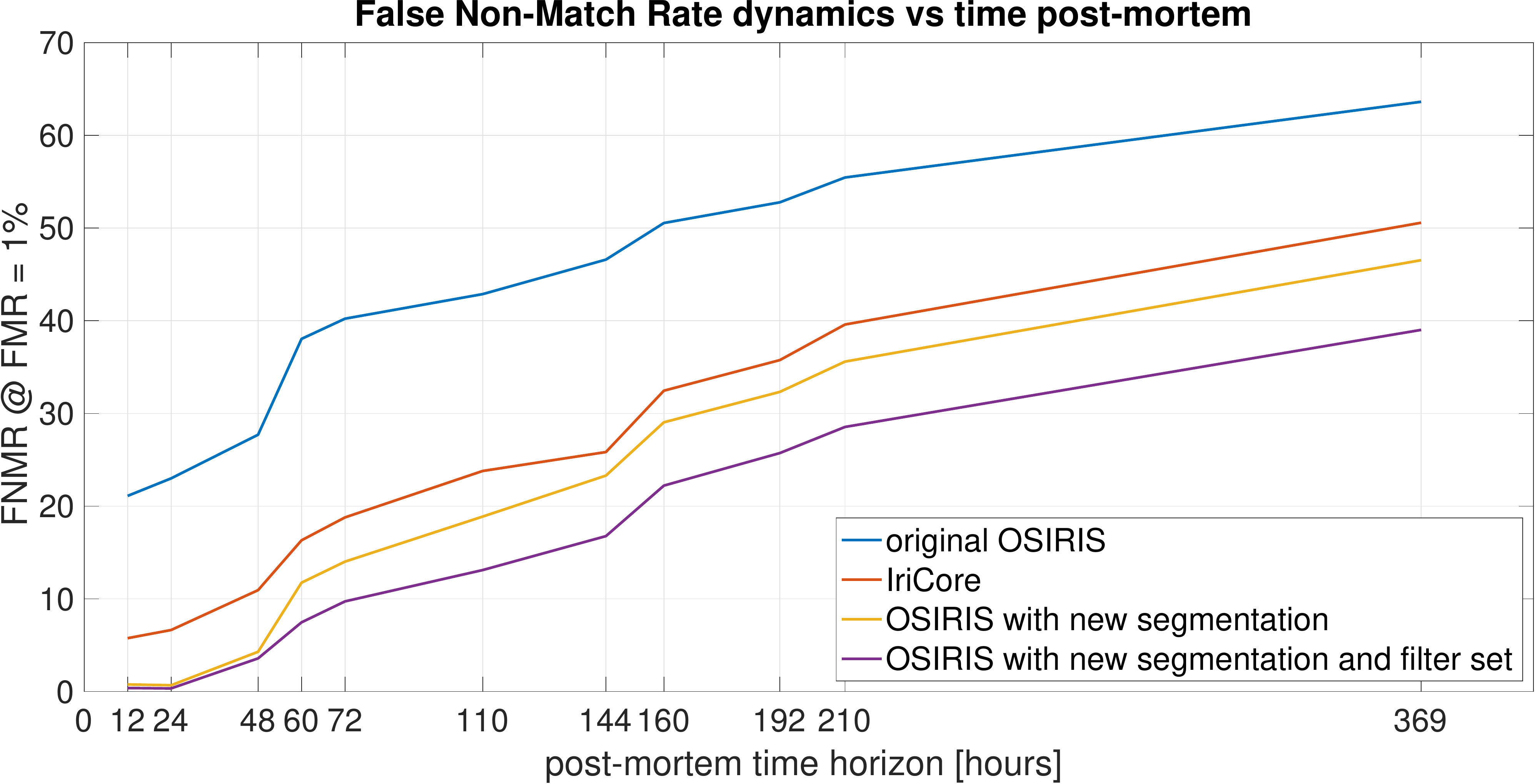}
\caption{Same as in Fig. \ref{fig:FNMR1_new}, but for a set FMR=1\%.}
\label{fig:FNMR1_new}
\end{figure*}

\begin{figure*}[h!]
\centering
\includegraphics[width=0.68\linewidth]{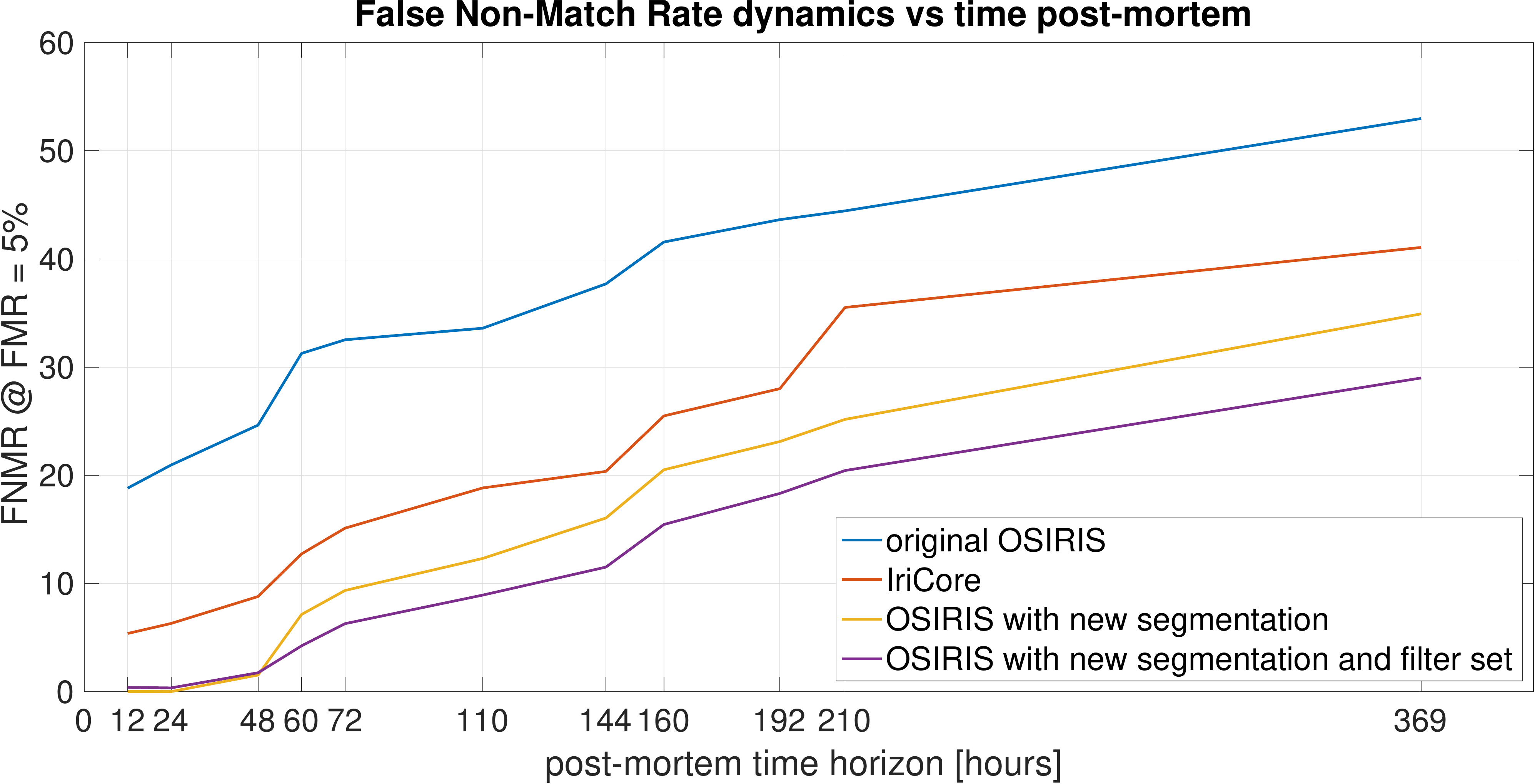}
\caption{Same as in Fig. \ref{fig:FNMR1_new}, but for a set FMR=5\%.}
\label{fig:FNMR5_new}
\end{figure*}

In addition to the ROCs, we have also calculated the False Non-Match Rate (FNMR) values at acceptance thresholds which allow the False Match Rate (FMR) values to stay below certain values, namely 0.1\%, 1\% and 5\%, Figs. \ref{fig:FNMR01_new}, \ref{fig:FNMR1_new}, \ref{fig:FNMR5_new}, respectively. This is done to reveal the dynamics of the FNMR as a function of post-mortem time interval, and therefore to know the chances for a false non-match as time since death progresses. We plot this dynamics for the two baseline methods: original OSIRIS and IriCore, as well as for the CNN-based segmentation modification, and the proposed iris representation, coupled with the new segmentation.

While acceptance thresholds allowing FMR of 5\% or even 1\% can be considered as very relaxed for large-scale iris recognition systems, such criteria make sense in a forensic scenario. In such, the goal of an automatic system is typically to aid a human expert by proposing a candidate list, while minimizing the chances of missing the correct hit. Therefore, allowing a higher False Match Rate will make it more likely for the correct hit to appear within the candidate list. 

Notably, for each moment during the increasing post-mortem sample capture time horizon, our proposed approach consistently offers an advantage over the other two algorithms, allowing to reach nearly perfect recognition accuracy for samples collected up to a day after a subject's death, as shown in Figs. \ref{fig:FNMR01_new}, \ref{fig:FNMR1_new}, and Fig. \ref{fig:FNMR5_new}. This difference in favor of our proposed solution is even larger when the acceptance threshold is relaxed to allow for 5\% False Match Rate -- in such a scenario, we can expect no false non-matches in the first 24 hours, and only approx. 1.5\% chance of a false non-match in the first 48 hours. The new method for iris feature representation, on the other hand, shows its greatest advantage in the acquisition horizons that are longer than 48 hours, being able to reduce the errors by as much as a third in the 60-hour and 72-hour capture moments.

To be fair, however, we need to stress that -- to our knowledge -- neither OSIRIS nor IriCore were designed to deal with post-mortem samples, so the lower performance of these methods is understandable. This only demonstrates that new iris methods capable to address post-mortem decomposition processes need to be designed, if we want to include iris into the basket of forensic identification tools.

\section{Conclusions}
\label{sec:Conclusions}
In this paper we have introduced the novel iris feature representation method, employing iris-specific image filters learnt directly from the data, and are thus optimized to be resilient against post-mortem changes affecting the eye during the increasing sample capture time since death. By finding the optimal combination of typical Gabor wavelet-based iris encoding with the new post-mortem-driven encoding we are able to reduce the recognition errors by as much as one third, significantly outperforming even the state-of-the-art commercial matcher. The source codes for the experiments, trained iris-specific filters, and the new segmentation model are made available along with the paper.

To our knowledge, it's the first post-mortem iris-specific and end-to-end recognition pipeline, open sourced and ready to be deployed in forensic applications. However, the segmentation model and the methodology of filter learning can be directly applied to other challenging cases in iris recognition, as eye with diseases, or irises acquired from infants or elderly people.


\begin{thebibliography}{10}\itemsep=-1pt

\bibitem{BostonPostMortem}
{A. Sansola}.
\newblock {Postmortem iris recognition and its application in human
  identification}, Master's Thesis, Boston University, 2015.

\bibitem{EwelinkaHandIWBF2018}
E.~Bartuzi, K.~Roszczewska, A.~Czajka, and A.~Pacut.
\newblock {Unconstrained Biometric Recognition Based on Thermal Hand Images}.
\newblock {\em International Workshop on Biometrics and Forensics (IWBF 2018)},
  2018.

\bibitem{bertinetto2016fullysiamesetracking}
L.~Bertinetto, J.~Valmadre, J.~F. Henriques, A.~Vedaldi, and P.~H. Torr.
\newblock Fully-convolutional siamese networks for object tracking.
\newblock {\em European Conference on Computer Vision (ECCV 2016)}, pages
  850--865, 2016.

\bibitem{BolmeBTAS2016}
D.~S. Bolme, R.~A. Tokola, C.~B. Boehnen, T.~B. Saul, K.~A. Sauerwein, and
  D.~W. Steadman.
\newblock Impact of environmental factors on biometric matching during human
  decomposition.
\newblock {\em IEEE 8th International Conference on Biometrics Theory,
  Applications and Systems (BTAS 2016), September 6-9, 2016, Buffalo, USA},
  2016.

\bibitem{BromleySiamense1993}
J.~Bromley, I.~Guyon, Y.~LeCun, E.~S\"{a}ckinger, and R.~Shah.
\newblock {Signature Verification Using a "Siamese" Time Delay Neural Network}.
\newblock {\em Proceedings of the 6th International Conference on Neural
  Information Processing Systems (NIPS 1993)}, pages 737--744, 1993.

\bibitem{CzajkaBSIFIrisDesc2019}
A.~Czajka, D.~Moreira, K.~W. Bowyer, and P.~J. Flynn.
\newblock Domain-specific human-inspired binarized statistical image features
  for iris recognition.
\newblock {\em IEEE Winter Conference on Applications of Computer Vision, (WACV
  2019), Hawaii, USA}, 2019.

\bibitem{SuperPointDeTone2018}
D.~{DeTone}, T.~{Malisiewicz}, and A.~{Rabinovich}.
\newblock {SuperPoint: Self-Supervised Interest Point Detection and
  Description}.
\newblock {\em IEEE/CVF Conference on Computer Vision and Pattern Recognition
  Workshops (CVPR'W 2018)}, 2018.

\bibitem{biosec}
J.~Fierrez, J.~Ortega-Garcia, D.~T. Toledano, and J.~Gonzalez-Rodriguez.
\newblock Biosec baseline corpus: A multimodal biometric database.
\newblock {\em Pattern Recognition}, 40(4):1389 -- 1392, 2007.

\bibitem{IriCore}
{IriTech Inc.}
\newblock {IriCore Software Developer's Manual}, version 3.6, 2013.

\bibitem{ADAM}
D.~P. Kingma and J.~Ba.
\newblock {Adam: A Method for Stochastic Optimization}.
\newblock 2014.

\bibitem{koch2015siamese}
G.~Koch, R.~Zemel, and R.~Salakhutdinov.
\newblock Siamese neural networks for one-shot image recognition.
\newblock {\em ICML Deep Learning Workshop}, 2, 2015.

\bibitem{MaheshwarySiameseResumes2018}
S.~Maheshwary and H.~Misra.
\newblock Matching resumes to jobs via deep siamese network.
\newblock {\em Companion Proceedings of the The Web Conference}, 2018.

\bibitem{bath}
D.~Monro, S.~Rakshit, and D.~Zhang.
\newblock {University of Bath, UK Iris Image Database}, 2009.

\bibitem{SiameseObjectCoSegmentation2018}
P.~Mukherjee, B.~Lall, and S.~Lattupally.
\newblock {Object cosegmentation using deep Siamese network}.
\newblock 2018.

\bibitem{VeriEye}
Neurotechnology.
\newblock {VeriEye SDK, version 4.3}, accessed: August 11, 2015.

\bibitem{OSIRIS}
N.~Othman, B.~Dorizzi, and S.~Garcia-Salicetti.
\newblock Osiris: An open source iris recognition software.
\newblock {\em Pattern Recognition Letters}, 82:124 -- 131, 2016.

\bibitem{ubiris}
H.~Proenca, S.~Filipe, R.~Santos, J.~Oliveira, and L.~Alexandre.
\newblock The {UBIRIS.v2}: A database of visible wavelength images captured
  on-the-move and at-a-distance.
\newblock {\em IEEE Transactions on Pattern Analysis and Machine Intelligence},
  32(8):1529--1535, 2010.

\bibitem{Sauerwein_JFO_2017}
K.~Sauerwein, T.~B. Saul, D.~W. Steadman, and C.~B. Boehnen.
\newblock The effect of decomposition on the efficacy of biometrics for
  positive identification.
\newblock {\em Journal of Forensic Sciences}, 62(6):1599--1602, 2017.

\bibitem{DeepDescSimoSerra2015}
E.~Simo-Serra, E.~Trulls, L.~Ferraz, I.~Kokkinos, P.~Fua, and F.~Moreno-Noguer.
\newblock {Discriminative Learning of Deep Convolutional Feature Point
  Descriptors}.
\newblock {\em IEEE International Conference on Computer Vision (ICCV 2015)},
  2015.

\bibitem{MIRLIN}
{Smart Sensors Ltd.}
\newblock {MIRLIN SDK}, version 2.23, 2013.

\bibitem{TrokielewiczPostMortemBTAS2016}
M.~Trokielewicz, A.~Czajka, and P.~Maciejewicz.
\newblock {Human Iris Recognition in Post-mortem Subjects: Study and Database}.
\newblock {\em 8th IEEE International Conference on Biometrics: Theory,
  Applications and Systems (BTAS 2016), September 6-9, 2016, Buffalo, USA},
  2016.

\bibitem{TrokielewiczPostMortemICB2016}
M.~Trokielewicz, A.~Czajka, and P.~Maciejewicz.
\newblock {Post-mortem Human Iris Recognition}.
\newblock {\em 9th IAPR International Conference on Biometrics (ICB 2016), June
  13-16, 2016, Halmstad, Sweden}, 2016.

\bibitem{TrokielewiczTIFS2018}
M.~Trokielewicz, A.~Czajka, and P.~Maciejewicz.
\newblock Iris recognition after death.
\newblock {\em IEEE Transactions on Information Forensics and Security},
  14(6):1501--1514, 2018.

\bibitem{TrokielewiczColdPAD_BTAS2018}
M.~Trokielewicz, A.~Czajka, and P.~Maciejewicz.
\newblock {Presentation Attack Detection for Cadaver Iris}.
\newblock {\em 9th IEEE International Conference on Biometrics: Theory,
  Applications and Systems (BTAS 2018), October 22-25, 2018, Los Angeles, USA},
  2018.

\bibitem{Trokielewicz_BTAS_2019}
M.~Trokielewicz, A.~Czajka, and P.~Maciejewicz.
\newblock {Perception of Image Features in Post-Mortem Iris Recognition: Humans
  vs Machines}.
\newblock {\em 10th IEEE International Conference on Biometrics: Theory,
  Applications and Systems (BTAS 2018), September 23-26, 2019, Tampa, USA},
  2019.

\bibitem{TrokielewiczIMAVIS2019arxiv}
M.~Trokielewicz, A.~Czajka, and P.~Maciejewicz.
\newblock {Post-mortem Iris Recognition with Deep-Learning-based Image
  Segmentation}.
\newblock {\em Arxiv preprint}, 2019.

\bibitem{VinyalsSiameseNIPS2016}
O.~Vinyals, C.~Blundell, T.~Lillicrap, k.~kavukcuoglu, and D.~Wierstra.
\newblock Matching networks for one shot learning.
\newblock volume~29, pages 3630--3638. 2016.

\bibitem{LIFT}
K.~M. Yi, E.~Trulls, V.~Lepetit, and P.~Fua.
\newblock {LIFT: Learned Invariant Feature Transform}.
\newblock {\em European Conference on Computer Vision (ECCV 2016)}, 2016.

\bibitem{ZagoruykoK15SiamesePatches}
S.~Zagoruyko and N.~Komodakis.
\newblock Learning to compare image patches via convolutional neural networks.
\newblock {\em Conference on Computer Vision and Pattern Recognition (CVPR
  2015)}, 2015.

\end{thebibliography}
\end{document}